\documentclass[runningheads]{llncs}

\usepackage[table,xcdraw]{xcolor}
\usepackage{graphicx}
\usepackage{array}
\usepackage{graphicx}
\usepackage{multirow}
\usepackage{ragged2e}
\usepackage{colortbl}
\usepackage{booktabs}
\usepackage{svg}
\usepackage[table]{xcolor}
\definecolor{mycolor1}{HTML}{7ED954}
\definecolor{mycolor2}{HTML}{99e17a}
\definecolor{mycolor3}{HTML}{bfecac}
\definecolor{mycolor4}{HTML}{d9f4cd}
\usepackage{eccv}

\usepackage{eccvabbrv}

\usepackage{graphicx}
\usepackage{booktabs}

\usepackage[accsupp]{axessibility}  

\usepackage{hyperref}

\usepackage{orcidlink}
\makeatletter
\newcommand{\printfnsymbol}[1]{%
  \textsuperscript{\@fnsymbol{#1}}%
}
\makeatother

\begin{document}

\title{Pixel2Perspective: A Benchmark for Multi-Modal and Multi-View Generalization in Vision-Language Models} 

\author{ Aditya Gandhamal\inst{2}\thanks{These authors contributed equally to this work.} \and Aniruddh Sikdar\inst{1}\printfnsymbol{1} \and Arya Kishor\inst{1} \and Suresh Sundaram\inst{1} } \authorrunning{A. Gandhamal et al.} \institute{ Department of Aerospace Engineering, Indian Institute of Science, Bengaluru, India  \and Kotak IISc AI-ML Centre, Indian Institute of Science, Bengaluru, India }

\maketitle

\begin{abstract}
  
Embodied AI systems, such as drones and autonomous vehicles, improve navigation by enhancing perception in environments containing unseen objects. While Visual Language Models (VLMs) offer strong image understanding, their real-world application is underexplored. For effective deployment, open-vocabulary semantic segmentation (OVSS) models, leveraging models like CLIP to assign pixel-level labels from text, must be robust to outdoor conditions, adverse weather, and varying viewpoints.
Despite their potential, these models encounter significant challenges in cross-domain generalization, hindering their practical use in embodied AI applications. In this work, we present Pixel2Perspective, a benchmark for multi-angle object localization and open-vocabulary segmentation across both aerial and ground perspectives. This benchmark supports thorough evaluation across different viewpoints and sensor modalities. We assess foundational models like CLIP and advanced OVSS models using a curated dataset that spans varied outdoor conditions, sim-to-real scenarios, and multiple sensors, including infrared, to support the development of robust and generalizable perception systems. We also present a novel framework, Multi-Source RGB–IR Uncertainty Quantification (MSRIQ), which jointly models aleatoric and epistemic uncertainties across RGB and infrared imagery. This approach supports robust evaluation of open-vocabulary segmentation models across diverse environmental conditions. Our work sets a robustness-driven benchmark, providing key insights and laying the foundation for future research in open-world embodied AI perception. The benchmark and project page are available at \href{https://aniruddhsikdar1.github.io/publications/pixel2perspective/}{this link}.

\end{abstract}

\section{Introduction}

Embodied AI systems, like autonomous vehicles and drones, depend on robust perception models for navigating dynamic environments, making cross-domain generalization crucial. Recent advances in Vision-Language Models (VLMs) have sparked interest in applying them to autonomous driving for their improved robustness \cite{sikdar2025picazo,gandhamal2025ov}. These models enhance multi-modal learning by integrating images and text, enabling better understanding of novel concepts, for example, CLIP, trained on 400 million image-text pairs. Open-vocabulary semantic segmentation (OVSS) has emerged as a promising approach for scene parsing, leveraging large-scale vision-language models pre-trained on massive internet datasets with language supervision, which demonstrate strong generalization \cite{shi2024lca, radford2021learning}. However, their reliance on image-level supervision limits performance on fine-grained tasks such as pixel-level segmentation \cite{zhou2022extract}.  Despite their potential, VLMs require thorough evaluation for fine-grained tasks, as cross-domain generalization challenges limit their effectiveness in embodied AI.

 \begin{figure}[]
    \centering
    \includegraphics[scale=0.15]{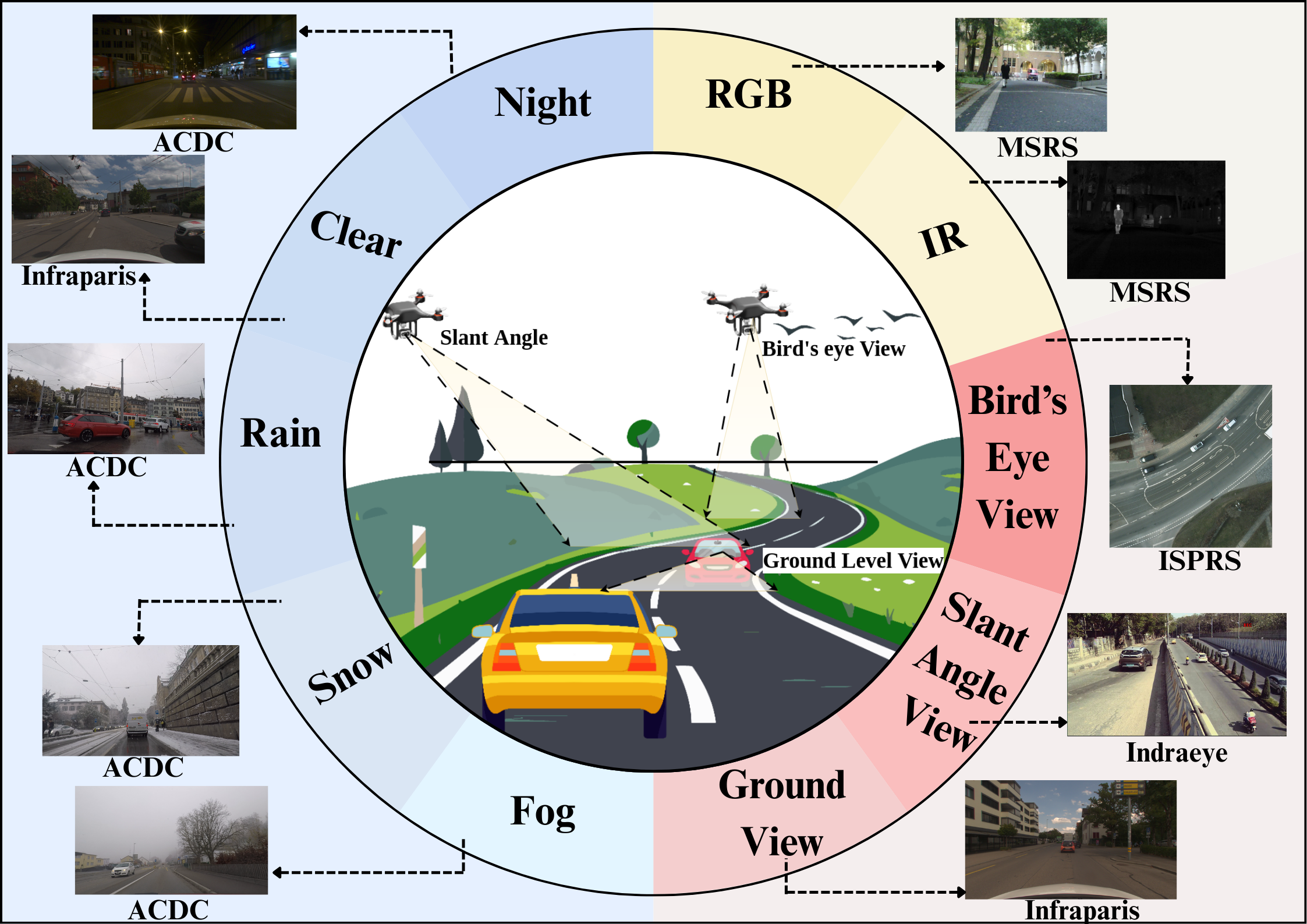}
    \caption{Viewpoint comparison in the Pixel2Perspective benchmark across three settings: (1) ground-level, (2) slant-angle, and (3) bird’s-eye view, including multiple modalities and weather conditions such as rain, snow, and fog.}
    \label{fig:p2p}
\end{figure}


Recent research has explored fine-tuning CLIP for fine-grained tasks and has explored improving OVSS models for segmentation using techniques such as mask proposal networks \cite{ghiasi2022scaling, yu2023convolutions} and cost aggregation methods \cite{cho2024cat}. However, pixel-level classification still struggles with unseen classes, particularly under significant domain shifts \cite{blumenstiel2024mess}. This limitation worsens as the number of unseen categories increases \cite{ding2022decoupling}, with open-vocabulary models being especially prone to overfitting during downstream task adaptation \cite{wortsman2022robust, kumar2022fine}.
Although OVSS models effectively capture open-set visual concepts for segmentation, their robustness remains largely unexplored.Their zero-shot generalization drops sharply under major category, distribution, or domain shifts, limiting open-vocabulary models’ real-world applicability \cite{zhang2024dept}.

Embodied AI systems require advanced perception capabilities to operate reliably in adverse weather and low-light conditions \cite{sikdar2025saga, chaudhuri2025domain}.  To address this, benchmarks such as the Multi-domain Evaluation of Semantic Segmentation (MESS) \cite{blumenstiel2024mess} have been developed to evaluate model performance across a wider range of complex, domain-specific datasets. However, leveraging foundation models like CLIP \cite{radford2021learning} for advanced perception presents three key challenges that remain unexplored in previous benchmarks: (1) fine-tuning on limited training dataset, for downstream segmentation tasks while preserving pretrained knowledge, (2) handling significant domain shifts arising from out-of-distribution data, and (3) accommodating substantial variations between seen and unseen classes, particularly those arising from differing viewing perspectives. These challenges are key to assessing model robustness across tasks. Reliable systems must work with visible (RGB) and infrared (IR) data, staying resilient to modality shifts and varying conditions like day-night cycles, fog, rain, and glare. 
Additionally, they must be capable of recognizing objects from multiple viewpoints, each offering unique semantic cues, despite often lacking the ability to choose the most informative perspectives \cite{seifi2021glimpse}. However, current standard benchmarks primarily target in-domain tasks, offering limited insight into how models perform across domains. This limitation restricts our understanding of zero-shot segmentation models’ ability to generalize to unseen domains and handle diverse sensors or vocabularies.

To tackle these challenges, we introduce Pixel2Perspective: a benchmark for multi-modal and multi-view generalization in vision-language models. It is an open-vocabulary RGB–Infrared segmentation benchmark designed for multi-angle object localization across drone, satellite, and ground-level perspectives, as shown in Figure \ref{fig:p2p}. We evaluate state-of-the-art models such as CLIP for fine-grained object localization, alongside leading open-vocabulary semantic segmentation (OVSS) models across 13 datasets. These datasets are carefully curated to span diverse object classes, sensor modalities, and out-of-distribution domains, with an emphasis on varied viewpoints to comprehensively assess generalization. We also introduce Multi-Source RGB–IR Uncertainty Quantification (MSRIQ), a benchmark framework that jointly models aleatoric and epistemic uncertainties across co-registered RGB and IR modalities, enabling robust evaluation of OVSS models in diverse conditions.

Our key contributions are as follows: (1) Pixel2Perspective Benchmark: We introduce Pixel2Perspective, a benchmark for evaluating state-of-the-art Vision-Language Models. Leveraging a quantitative taxonomy of diverse segmentation datasets, it rigorously assesses out-of-domain object localization and  OVSS performance across ground-level, aerial, and bird’s-eye perspectives, under varied weather conditions, and across multiple modalities including RGB and IR. (2) Systematic Evaluation of VLMs: We conduct a comprehensive evaluation of Vision-Language Models on object localization and open-vocabulary semantic segmentation, offering an in-depth analysis across diverse settings. (3) Multi-Source RGB–IR Uncertainty Quantification (MSRIQ): To handle uncertainties from both RGB and IR sensors, we propose MSRIQ, a framework that jointly models aleatoric and epistemic uncertainties across co-registered RGB and infrared modalities.

 \begin{figure*}[]
    \centering
    \includegraphics[scale=0.25]{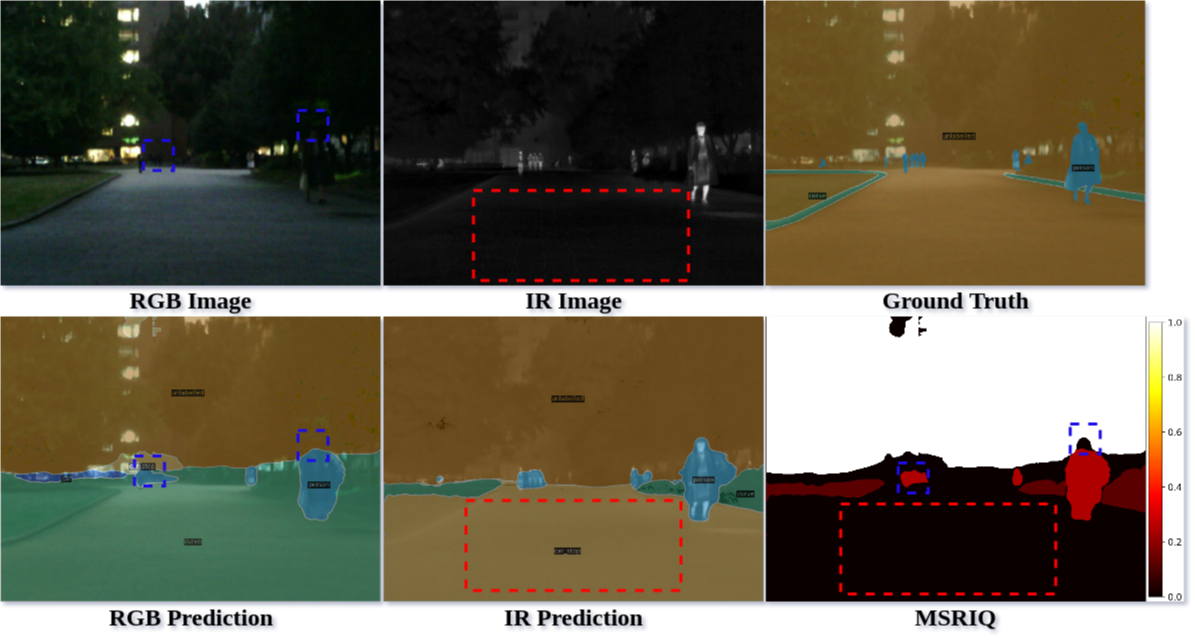}
    \caption{Qualitative comparison of CAT-Seg outputs on RGB and IR images from the MSRS dataset, with MSRIQ illustrating the proposed uncertainty metric to clearly indicate regions of maximum uncertainty.}
    \label{fig:MSRIQ}
\end{figure*}

\begin{figure}[]
    \centering
    \includegraphics[scale=0.16]{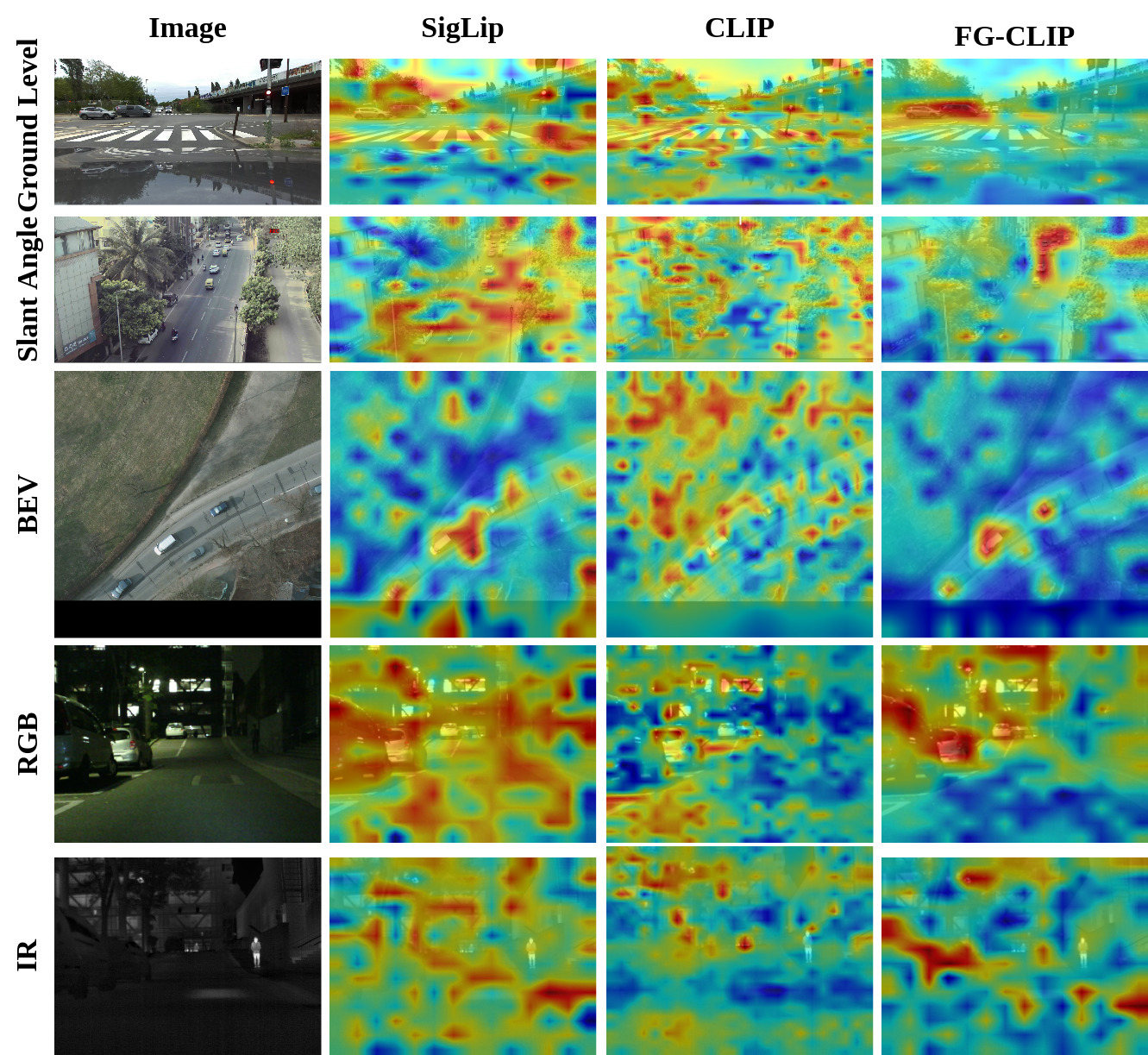}
    \caption{As shown, we visualize the logits of the vision and text embeddings from the ViT-B CLIP encoder to assess object localization capabilities. The heatmaps for CLIP, SigCLIP, and FC-CLIP highlight regions most correlated with the text description: “A photo of a \textit{CAR} in the scene.”}
    \label{fig:clip-gradcam}
\end{figure}

\section{Related Works}

\textbf{Open-vocabulary semantic segmentation} Large-scale pre-training has revolutionized computer vision by training models on image-text pairs, enabling representation of visual and textual semantics in a shared space. Open-vocabulary semantic segmentation (OVSS) aims to bridge the gap between image-level and pixel-level understanding by enabling segmentation of unseen classes. OVSS models are generally categorized into two main types: (i) mask-proposal networks \cite{ghiasi2022scaling, ding2022open, xu2023open, xu2022simple, xu2023side, yu2023convolutions}, and (ii) cost-aggregation networks \cite{cho2024cat}. Mask-proposal approaches generate class-agnostic region proposals, which are then semantically aligned with CLIP-derived textual embeddings to facilitate open-vocabulary classification at the pixel level. For instance, OpenSeg \cite{ghiasi2022scaling} leverages local image regions to form proposals, whereas models like ZegFormer \cite{ding2022decoupling} and ZSseg \cite{xu2022simple} adopt a two-stage architecture to address this task.   CAT-Seg \cite{cho2024cat} enhances robustness by using similarity scores to combine CLIP’s image and text embeddings into a cost volume, rather than relying on direct embeddings. OVSS models are commonly evaluated on standard datasets like ADE20K \cite{zhou2019semantic} and Pascal VOC \cite{everingham2015pascal}, which primarily feature everyday, in-domain images. However, such evaluations fail to capture real-world conditions with inherent domain shifts and environmental diversity.  \\
\textbf{Related Benchmarks} OVSS models are typically evaluated on datasets composed exclusively of in-domain images \cite{zhou2019semantic,everingham2015pascal}, which are considered the de facto standard for model evaluation \cite{ghiasi2022scaling, liang2023open}. Few works have explored broader dataset diversity. The Segmentation in the Wild (SegInW) benchmark \cite{zou2023generalized} includes 25 datasets, however, most of them still feature everyday, in-domain images, with only two notable exceptions.  Robust benchmarks are crucial for advancing the vision community, providing standardized ways to evaluate performance across diverse tasks \cite{paul2022vision}. Inspired by HELM \cite{liang2022holistic} for LLMs, the MESS benchmark \cite{blumenstiel2024mess} was introduced to assess OVSS models across domains such as medicine, engineering, earth observation, biology, and agriculture. However, a key limitation lies in the relatively low number of classes within these datasets, resulting in high random mIoU scores.  To the best of our knowledge, open-vocabulary semantic segmentation (OVSS) models have not been systematically evaluated on datasets that account for sensor shifts and variations in look-angle, as commonly encountered in drones and autonomous vehicles. This work presents the first comprehensive effort to address these challenges, aiming to provide new insights and advance the robustness of OVSS models in practical settings.


\begin{table*}[ht]

\centering
\caption{Comparison of state-of-the-art OVSS methods trained on COCO-Stuff and evaluated on the RGB modality of the Pixel2Perspective Benchmark.}
\resizebox{\textwidth}{!}{
\begin{tabular}{c| ccc ccc ccc cccc}
\hline

\multirow{2}{*}{Method} 
    & \multicolumn{3}{c}{\cellcolor {mycolor4!50 }General}
    & \multicolumn{3}{c}{\cellcolor{mycolor3!50}Autonomous}
    & \multicolumn{3}{c}{\cellcolor {mycolor2!50}Slant-angle}
    & \multicolumn{4}{c}{\cellcolor{mycolor1!50}BEV} \\
\cline{2-14}
    & MSRS & MVSeg & SemanticRT
    & ACDC & MFNet & Infraparis
    & CART & UAVid & Indraeye
    & ISPRS & Worldfloods & LoveDA & OpenEarthMap \\
\hline
Zegformer   & 12.48 & 11.6 & 9.55   & 31.81   & 10.41   & 24.22  & 17.18   & 17.85   & 12.92   & 14.56   & 13.6   & 7.68    & 10.10 \\
SAN         & 20.35 & 21.96 & 12.56 & 35.49   & 21.05   & 31.04  & 37.36   & 41.65   & 7.84    & 37.56   & 30.10  & 19.53   & 22.52 \\
OVSeg       & 19.36 & 14.87 & 14.25 & 38.37   & 18.32   & 25.58  & 31.67   & 33.69   & 13.4    & 23.56   & 30.14  & 7.42    & 12.37 \\
CAT-Seg     & 26.63 & 22.27 & 16.16 & 39.09   & 28.87   & 34.26  & 39.66   & 41.55   & 12.84   & 50.1    & 38.25  & 11.61   & 22.90 \\
\hline
Zegformer-L & & & & & & & & & & & & & \\
SAN-L       & 17.07 & 23.90 & 16.62 & 43.99   & 17.22   & 36.17  & 39.39   & 43.41   & 10.56   & 52.06   & 45.93  & 20.08   & 23.79 \\
OVSeg-L &19.98	&19.58	&14.01	&39.97	&19.21	&24.82	&35.21	&38.79	&16.44	&35.46	&22.87	&19.20	&18.53   \\
CAT-Seg-L   & 20.77 & 26.47 & 17.44 & 47.47   & 22.20   & 38.35  & 39.85   & 41.90   & 16.22   & 50.92   & 39.54  & 40.87   & 26.61 \\
\hline
\end{tabular}
}
\label{Table:mscoco}
\end{table*}

\begin{table*}
\centering

\caption{Comparison of state-of-the-art OVSS methods trained on COCO-Stuff and evaluated on the IR modality of the Pixel2Perspective Benchmark.}
\resizebox{\textwidth}{!}{
\begin{tabular}{c| ccc cc cc cc} 
\hline
\multirow{2}{*}{Method} & \multicolumn{3}{c}{{\cellcolor[rgb]{0.835,0.89,0.941}}General} & \multicolumn{2}{c}{{\cellcolor[rgb]{0.757,0.835,0.941}}Autonomous} & \multicolumn{2}{c}{{\cellcolor[rgb]{0.706,0.796,0.941}}Slant-angle} & \multicolumn{2}{c}{{\cellcolor[rgb]{0.659,0.71,0.878}}BEV}  \\ 
\cline{2-10}
                        & MSRS    & MVSeg  & SemanticRT                                  & MFNet   & Infraparis                                               & CART    & Indraeye                                                  & ISPRS (irrg) & Worldfloods                                  \\ 
\hline
Zegformer               & 7.38    & 3.4    & 3.65                                        & 4.44    & 5.59                                                     & 3.34    & 3.21                                                      & 12.36        & 12.3                                         \\
SAN                     & 19.93   & 11.93  & 11.2894                                     & 18.1561 & 12.7672                                                  & 25.35   & 4.91                                                      & 45.0565      & 31.7526                                      \\
OVSeg                   & 13.91   & 5.29   & 9.99                                        & 12.58   & 10.19                                                    & 16.46   & 6.2                                                       & 17.52        & 37.36                                        \\
CAT-Seg                 & 17.85   & 12.62  & 11.59                                       & 21.73   & 13.61                                                    & 20.02   & 6.7                                                       & 45.36        & 35.72                                        \\ 
\hline
SAN-L                   & 12.7325 & 14.812 & 12.3079                                     & 14.3064 & 16.0332                                                  & 21.1193 & 7.5556                                                    & 51.4542      & 48.2444                                      \\
OVSeg-L                 & 15.53   & 9.01   & 13.4947                                     & 10.9737 & 11.3728                                                  & 20.7    & 12.48                                                     & 31.0294      & 31.4848                                      \\
CAT-Seg-L               & 19.44   & 15.32  & 17.6009                                     & 21.9945 & 15.6032                                                  & 25.84   & 11.74                                                     & 50.9261      & 44.4439                                      \\
\hline
\end{tabular}
}
\label{table:mscoco-ir}
\end{table*}

\section{Pixel2Perspective Benchmark}

\textbf{Problem Formulation} To evaluate the visual understanding of VLMs across different viewpoints and modalities, we focus on two key tasks: \textit{object localization} and \textit{Open vocabulary semantic segmentation (OVSS)}. We define object localization as the task of determining an object’s location based on a given textual prompt, using both image and text encoders. For the OVSS task, let \(I\) represent an image associated with a set of classes \(C = \{C_{1}, C_{2}, \dots, C_{N}\}\), where each class \(C_i\) is defined by a textual label. OVSS models aim to assign a class label \(C_i\) to each pixel in image \(I\), where the set of candidate classes \(N\) can vary at inference time and may include novel categories not seen during training.

\subsection{Dataset Taxonomy}

Robustness challenges in real-world deployment of perception models are frequently overlooked in existing research. To tackle this, we introduce Pixel2Perspective-Benchmark, an open-vocabulary RGB–IR benchmark for multi-angle segmentation across drones, satellites, and ground perspectives. 
The benchmark focuses on two key aspects, where systematic evaluation is essential: (1) assessing multi-modal backbones to measure the fine-grained capabilities of encoders, given their crucial role in retaining essential world knowledge, and (2) evaluating OVSS methods for fine-grained segmentation tasks.

As part of creating this benchmark, we reviewed a large collection of open-source datasets and filtered them based on three exclusion criteria: (1) viewing angle, (2) sensor modality, and (3) environmental conditions (day, night, fog, rain). Guided by the proposed taxonomy, we grouped the datasets into three primary viewpoint categories: (1) ground-level, (2) slant-angle, and (3) bird’s-eye view. This categorization facilitates evaluating a model’s ability to segment objects across significantly different viewpoints. We also assess performance using both RGB and IR modalities to ensure robustness under varied environmental and weather conditions. Datasets with unsuitable resolution or limited to binary labels were excluded from the benchmark.

Guided by the developed taxonomy, we carefully curated a representative set of datasets to ensure the benchmark is both informative and reproducible. Through iterative refinement, we identified 13 most relevant datasets for each viewpoint category and assessed model performance across diverse domain and sensor variations.
For ground-level perspectives, we selected the MSRS \cite{tang2022piafusion}, MVSeg \cite{ji2023multispectral}, SemanticRT \cite{ji2023semanticrt}, MFNet \cite{yuan2024unirgb}, and InfraParis \cite{franchi2024infraparis} datasets, all of which feature IR modalities. To evaluate performance under adverse weather conditions, we used the ACDC dataset \cite{sakaridis2021acdc}. To represent varying slant-angle perspectives, we incorporated the UAVid \cite{lyu2020uavid}, IndraEye \cite{sikdar2025saga}, and CART \cite{lee2024caltech} datasets, with IndraEye and CART also providing IR imagery. For bird’s-eye views, we selected the ISPRS dataset \cite{swissphoto2012isprs}, Worldfloods \cite{portales2023global}, OpenEarthMap \cite{xia2023openearthmap}, and LoveDA \cite{wang2021loveda} datasets. \textit{(A detailed table of selected datasets, instance counts, and benchmark comparisons is provided in the supplementary material.)}

\subsection{Fine-Grained Assessment Settings}

For object localization, we employ CLIP and other CLIP-based models, tailored for fine-grained tasks, in a zero-shot setting.
For the OVSS task, we define multiple evaluation settings: (1) training on COCO-Stuff \cite{caesar2018coco}, (2) training on the slant-angle  IndraEye (RGB) dataset, (3) training on the ground-level MSRS and ACDC datasets, and (4) Training on bird’s-eye view OpenEarthMap, covering both RGB and sim-to-real setups. These settings assess the models’ ability to generalize across different viewpoints by training on a specific angle and validating across all perspectives. Setting (1) represents the standard evaluation for fine-grained OVSS tasks and evaluates the effectiveness of VLMs in leveraging pre-training knowledge across diverse look angles.
Other Settings introduce additional challenges with limited training samples. Datasets were selected based on official test splits, high-quality annotations, and sufficient image resolution, and were used in compliance with their licenses without ethical concerns. This benchmark is well-suited for evaluating model robustness, testing performance across multiple look angles, and addressing challenges such as weather variations, sensor degradation, complex environments, and viewpoint-dependent semantic changes.

\subsection{Multi-Source RGB–IR Uncertainty Quantification}

Uncertainty in perception can be categorized into two types: aleatoric uncertainty, which arises from noise inherent in observations such as sensor limitations, and epistemic uncertainty, which originates from the model itself \cite{kendall2017uncertainties}. As shown in Fig. \ref{fig:MSRIQ}, aleatoric uncertainty is evident in the RGB image when the sensor struggles to detect a person in low-light conditions, resulting in uncertain OVSS predictions. Although the IR sensor can detect the person through thermal radiation, the model’s predictions remain uncertain due to the domain shift from RGB to IR caused by RGB-based pre-training, highlighting epistemic uncertainty. This emphasizes the need to assess such models on IR data to quantify cross-modality uncertainty. To address uncertainties from both RGB and IR sensors, we propose Multi-Source RGB–IR Uncertainty Quantification (MSRIQ), a framework that jointly models aleatoric and epistemic uncertainties across co-registered RGB and infrared modalities to evaluate segmentation performance. As shown in the Fig.  \ref{fig:MSRIQ}, the Intersection over Union (IoU) for the RGB image is computed as:

\begin{equation}
IoU = \frac{|P_{RGB} \cap GroundTruth|}{|P_{RGB} \cup GroundTruth|}
\end{equation}

where $P_{RGB}$ and $P_{IR}$  are the OVSS model’s predictions for the RGB and IR images, respectively, compared against the \textit{GroundTruth} to compute the IoU. We define Multi-Source RGB–IR Uncertainty Quantification (MSRIQ) as:

\begin{equation}
IoU_{MSRIQ} = \frac{|P_{RGB} \cap P_{IR}|}{|P_{RGB} \cup P_{IR}|}
\end{equation}

This metric quantifies the pixel-level spatial overlap between the RGB and IR predictions. An MSRIQ value close to 0 indicates little or no overlap between the predictions. This approach enables a more robust evaluation of open-vocabulary segmentation models across diverse and challenging environmental conditions. For scenarios where the pixels are not co-registered, we propose using Maximum Mean Discrepancy (MMD) to compare the distributions.

\section{Experiment Setup}

\textbf{Benchmarking Vision-Language Models}  For object localization, we qualitatively assess GradCAM visualizations for foundational models such as CLIP \cite{radford2021learning}, SigCLIP \cite{zhai2023sigmoid}, and FC-CLIP \cite{xie2025fg}. For the OVSS task, we evaluate state-of-the-art models including OVSeg \cite{liang2023open}, Zegformer \cite{ding2022decoupling}, SAN \cite{xu2023side}, and CAT-Seg \cite{cho2024cat}, which employ diverse architectural designs, ranging from two-stage mask-based and one-stage mask-based to pixel-based approaches. These models differ in architecture and parameter scale, are trained on a variety of vision-language tasks, and demonstrate strong visual understanding. All evaluations are performed using the mean intersection over union (mIoU) metric, based on their official implementations.

\subsection{Experimental Evaluation and Insights}


 \begin{figure}[]
    \centering
    \includegraphics[scale=0.15]{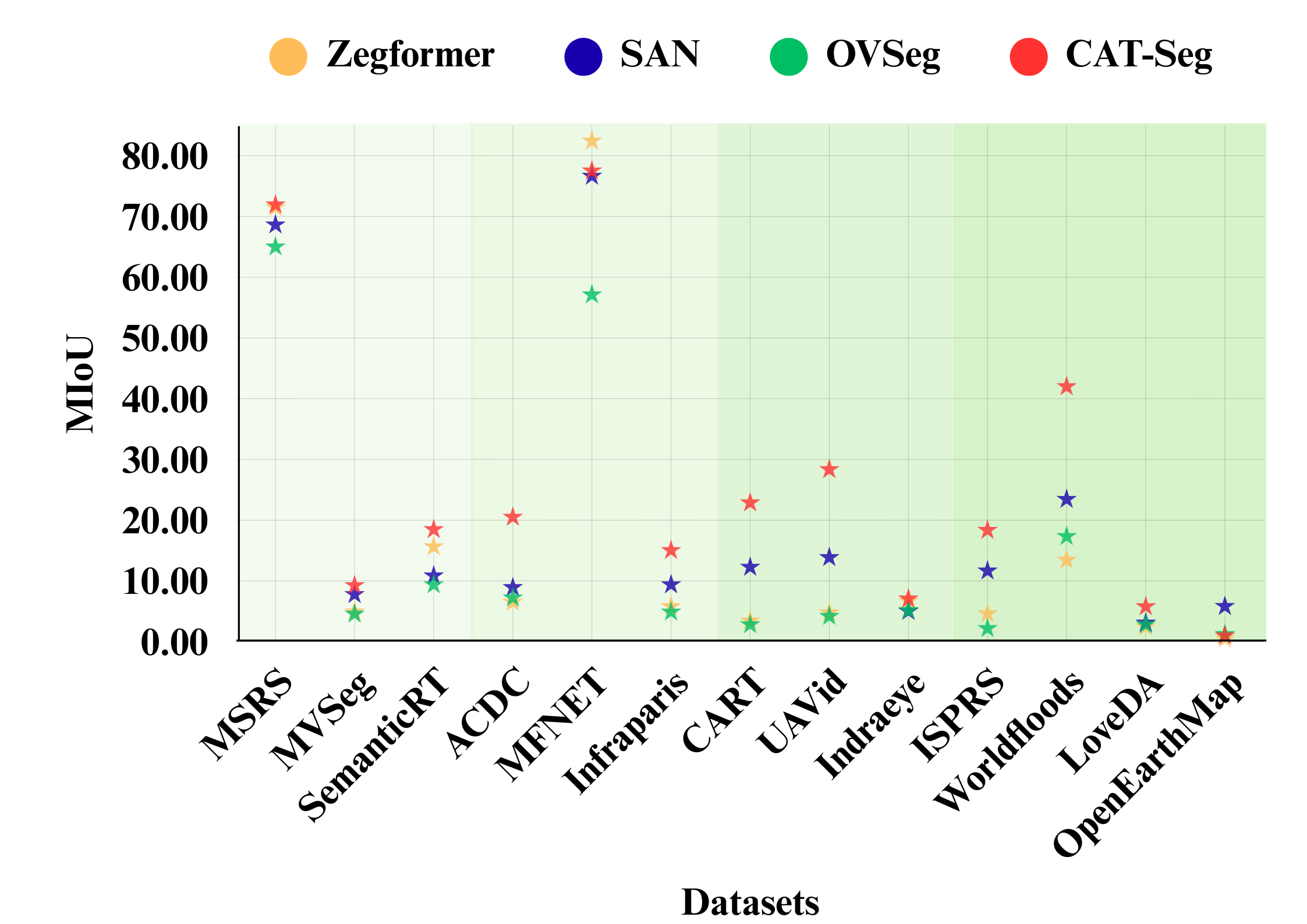}
    \caption{Scatter plot of OVSS models trained on MSRS (RGB) and evaluated on the Pixel2Perspective Benchmark (RGB).}
    \label{fig:sc_1}
\end{figure}

\begin{figure}[]
    \centering
    \includegraphics[scale=0.14]{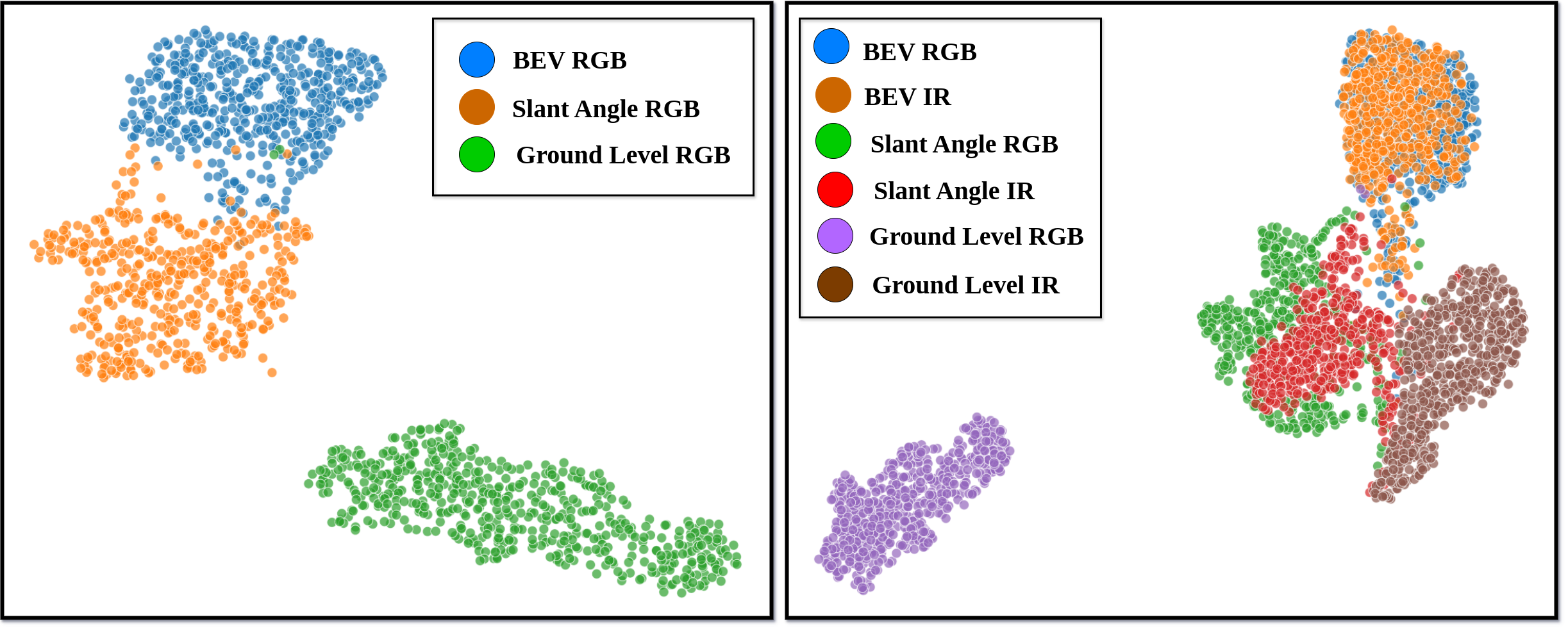}
    \caption{UMAP visualization of viewpoints and modality gaps. Colored points represent feature distributions from different perspectives (Ground Level, Slant Angle, BEV) and modalities (RGB, IR). The left plot (RGB) shows clearly separated clusters, highlighting strong viewpoint-specific domain gaps. The right plot (RGB + IR) shows overlapping and dispersed clusters, indicating both inter-perspective and cross-modal domain shifts. These visualizations demonstrate that embeddings are highly dependent on viewpoint and modality, which limits generalization.}
    \label{fig:umap}
\end{figure}

\begin{figure}[]
    \centering
    \includegraphics[scale=0.15]{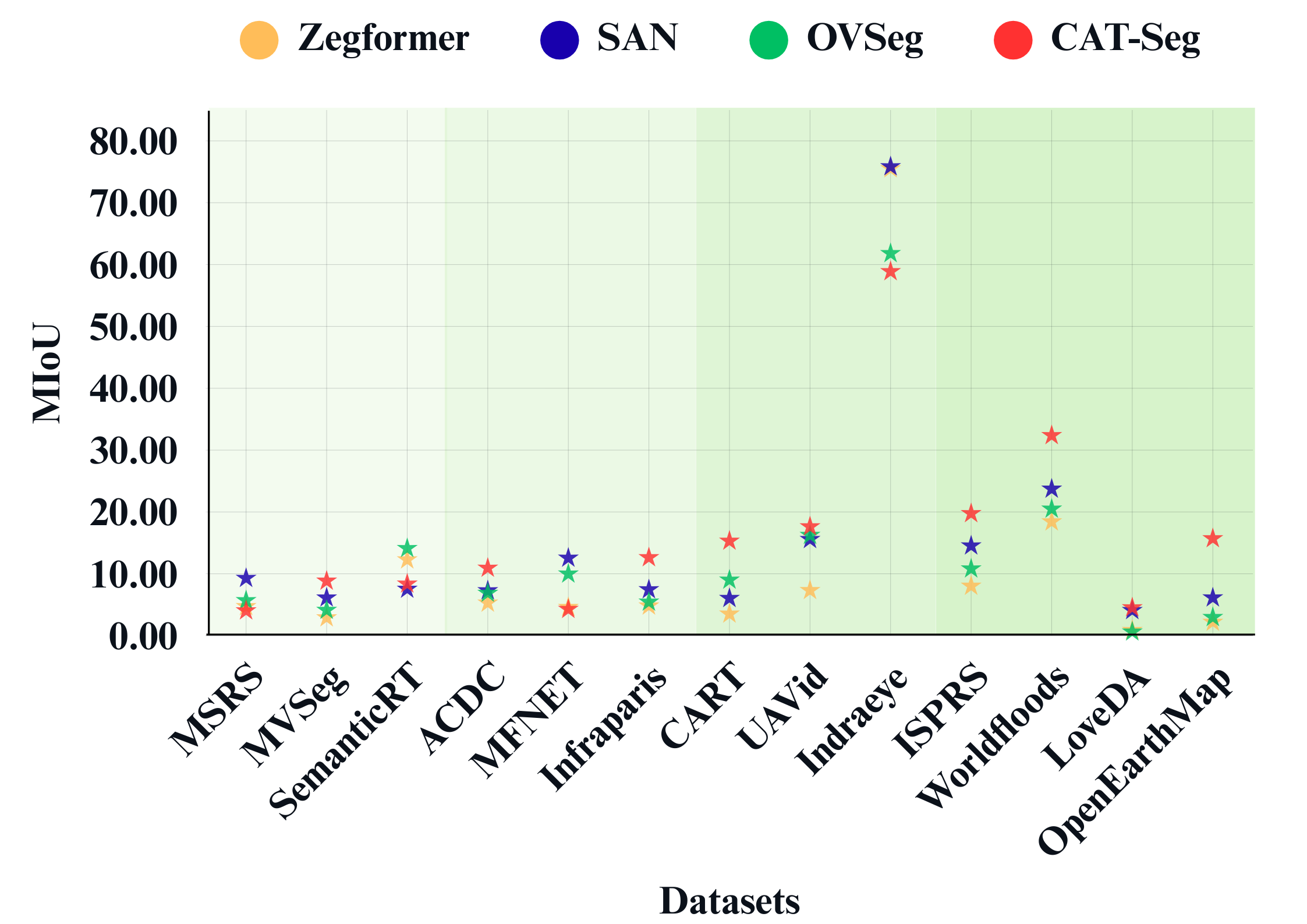}
    \caption{Scatter plot of OVSS models trained on Indraeye (RGB) and evaluated on the Pixel2Perspective Benchmark (RGB).}
    \label{fig:sc_3}
\end{figure}

\begin{figure}[]
    \centering
    \includegraphics[scale=0.25]{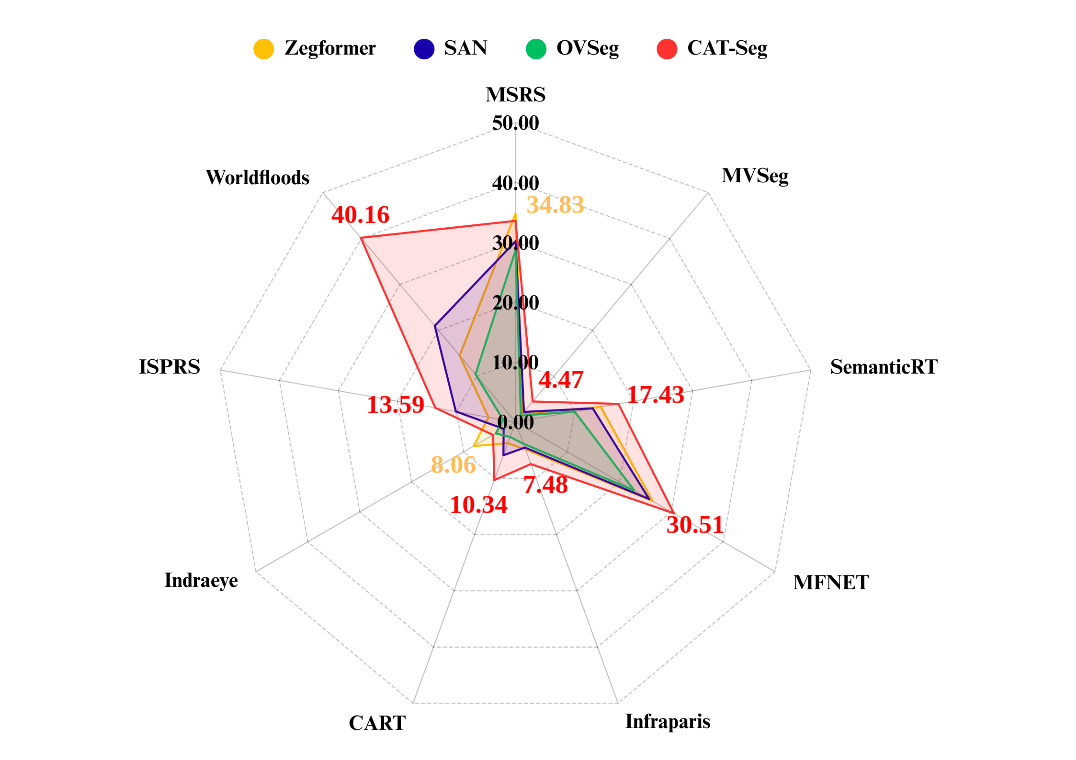}
    \caption{Radar plot of OVSS models trained on MSRS (RGB) and evaluated on the Pixel2Perspective Benchmark (IR).}
    \label{fig:radar1}
\end{figure}

\begin{figure}[]
    \centering
    \includegraphics[scale=0.25]{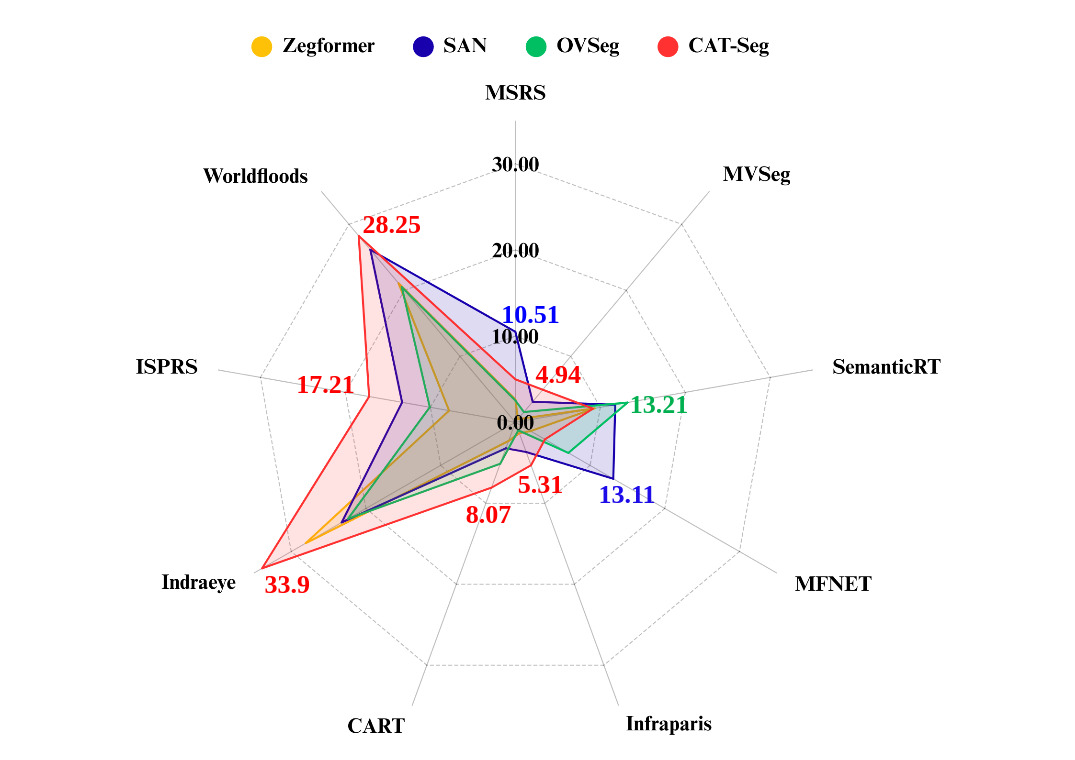}
    \caption{Radar plot of OVSS models trained on IndraEye (RGB) and evaluated on the Pixel2Perspective Benchmark (IR).
    }
    \label{fig:radar3}
\end{figure}

\textbf{Object Localization} We qualitatively evaluate GradCAM visualizations for CLIP \cite{radford2021learning}, SigCLIP \cite{zhai2023sigmoid}, and FC-CLIP \cite{xie2025fg}. As shown in Fig. \ref{fig:clip-gradcam}, FC-CLIP performs better across multiple viewpoints but struggles with night views and IR sensors. Similar conclusions can be drawn from the GradCAM visualizations, indicating that CAT-Seg (OVSS model) struggles to generalize across viewpoints even after fine-tuning on specific perspectives (see supplementary material for results).\\
\textbf{Multi-domain Evaluation}  Pixel2Perspective's real-world benchmarking is particularly effective for assessing the robustness under conditions that closely resemble those found in real-world deployments. Table \ref{Table:mscoco} shows a comparison of OVSS models on the Pixel2Perspective Benchmark, trained on the COCO-Stuff dataset and evaluated using only the RGB modality. Table \ref{table:mscoco-ir} presents the evaluation on the IR modality. Most OVSS models are typically evaluated on in-domain datasets. As noted, the CAT-Seg model performs better than other models, when trained on COCO-Stuff \\
\textbf{Robustness for modalities beyond the visible spectrum} Figures \ref{fig:radar1}, and \ref{fig:radar3} depict the performance of OVSS models trained on various datasets and evaluated on IR modalities. As seen, encoders trained primarily on RGB data exhibit limited robustness to other sensors, a limitation apparent in both CLIP-based encoders and OVSS models. Even when trained on the MSRS RGB dataset, performance on co-registered IR pairs drops by over 50\%, and for noisy sensors, such as those in the MVSeg dataset, performance declines further. Additionally, the benchmark empirically demonstrates that current OVSS models struggle to segment objects in infrared imagery, highlighting sensor-shift as a critical challenge. For CAT-Seg model, the proposed MSRIQ metric (Fig. \ref{fig:MSRIQ}) effectively highlights areas of highest model uncertainty. Results for models trained on ACDC and Openearthmap are provided in the supplementary material. \\
\textbf{Evaluation under diverse viewing angles} Models trained on diverse viewpoints are evaluated using the Pixel2Perspective Benchmark. Figure \ref{fig:umap} illustrates the domain gaps between different viewpoints. Scatter plot \ref{fig:sc_1} shows a significant drop in performance for datasets with differing look-angles, indicating that testing VLMs on a single viewpoint is insufficient. For instance, as seen in Fig. \ref{fig:sc_3}, models trained on slant-angle datasets like IndraEye generalize poorly to ground-level datasets such as MSRS and MVSeg, emphasizing the reliance of OVSS models on pre-training for generalization across unseen viewpoints. These results emphasize the importance of improving VLMs to enable reliable perception in embodied AI systems.

We also compute Maximum Mean Discrepancy (MMD) distance matrices between domain feature distributions under two settings, following fine-tuning on a perspective-specific dataset (see supplementary material). Compared to the zero-shot scenario, fine-tuning consistently produces higher MMD scores, indicating increased divergence of embeddings across domains. This illustrates a trade-off: while fine-tuning on a single perspective enhances within-domain alignment, it diminishes the model’s ability to generalize to other viewpoints. Although these classes were likely present during the pre-training of foundation models like CLIP, their semantics are biased toward certain perspectives, making it challenging for the models to recognize objects in different viewpoints.

\begin{figure}[]
    \centering
    \includegraphics[scale=0.15]{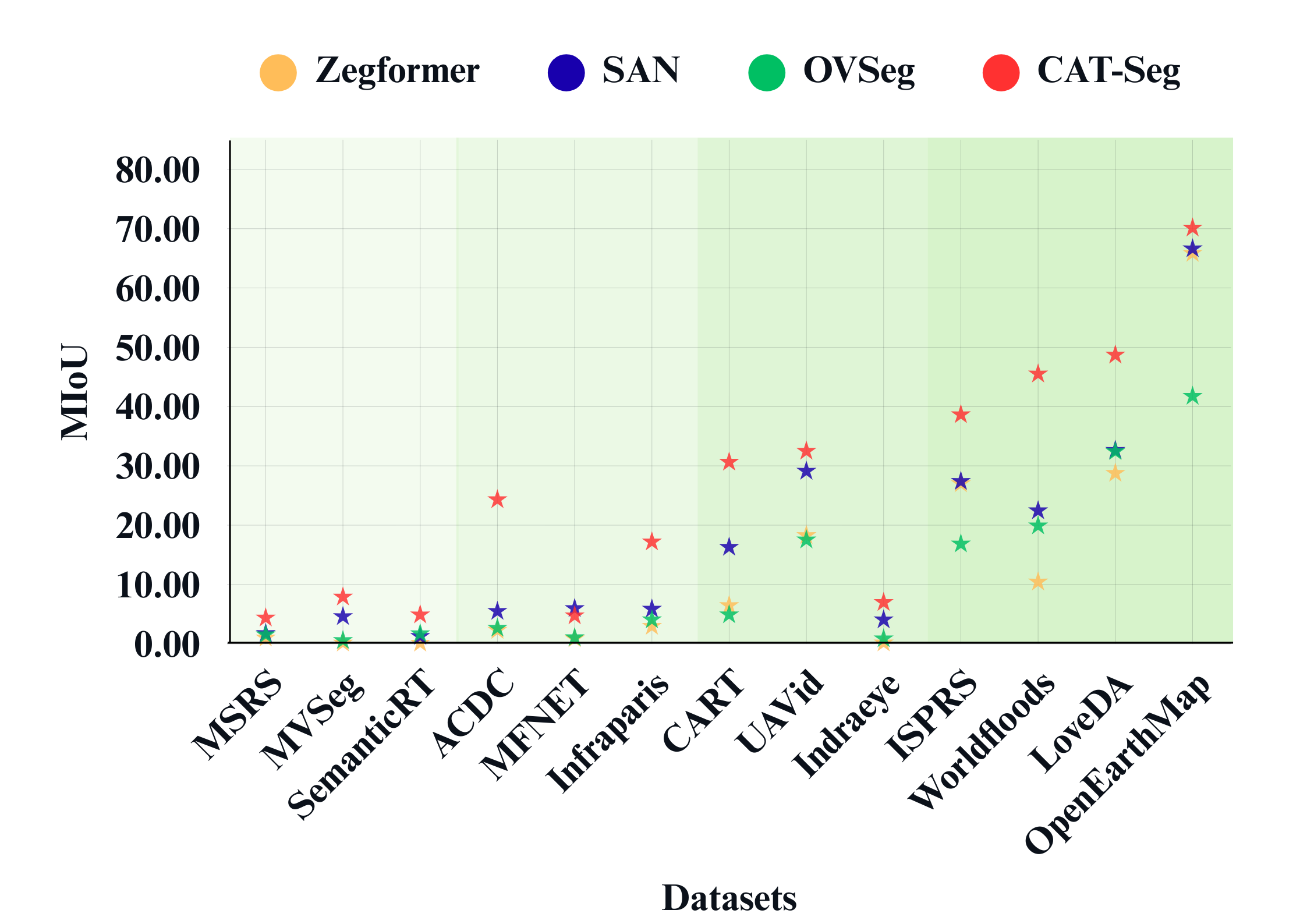}
    \caption{Scatter plot of OVSS models trained on Openearthmap (RGB) and evaluated on the Pixel2Perspective Benchmark (RGB).}
    \label{fig:sc_4}
\end{figure}

 \begin{figure}[]
    \centering
    \includegraphics[scale=0.25]{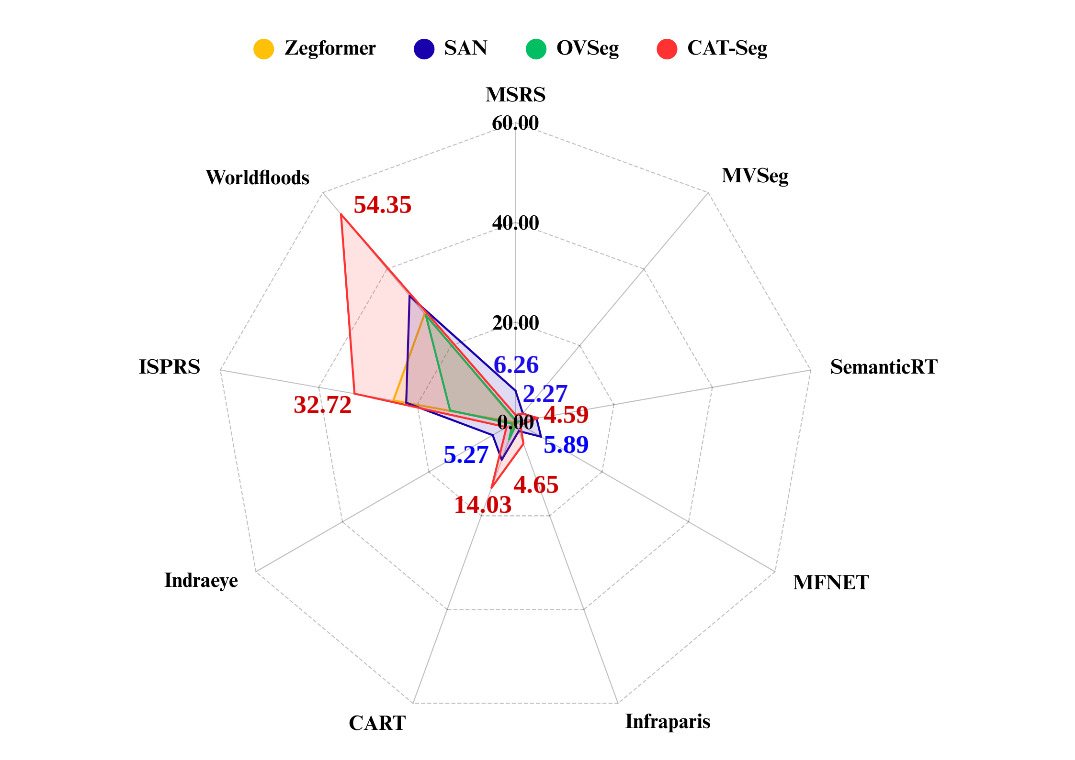}
    \caption{Radar plot of OVSS models trained on Openearthmap (RGB) and evaluated on the Pixel2Perspective Benchmark (IR).
    }
    \label{fig:radar4}
\end{figure}

\section{Conclusion}

This paper presents Pixel2Perspective, a benchmark for evaluating OVSS model robustness across multiple views and sensor modalities, a crucial step toward real-world deployment. Through systematic evaluation of state-of-the-art models in object localization and OVSS tasks, we reveal a substantial gap between current model capabilities and real-world demands, highlighting the need to enhance semantic understanding. Our robustness evaluation framework, along with the proposed Multi-Source RGB–IR Uncertainty Quantification (MSRIQ), provides valuable tools for developing more resilient models. By addressing these challenges, the benchmark provides key insights for embodied AI systems aiming to enhance perception and autonomous navigation in ground vehicles and drones. It also lays the groundwork for future research into why generalization across viewpoints remains difficult in practical settings.

%
%
\bibliographystyle{splncs04}
\bibliography{main}

@String(AAAI  = {AAAI})

@String(AAAI = {AAAI})

@inproceedings{zhang2024dept,
  title={Dept: Decoupled prompt tuning},
  author={Zhang, Ji and Wu, Shihan and Gao, Lianli and Shen, Heng Tao and Song, Jingkuan},
  booktitle={Proceedings of the IEEE/CVF Conference on Computer Vision and Pattern Recognition},
  pages={12924--12933},
  year={2024}
}

@inproceedings{ghiasi2022scaling,
  title={Scaling open-vocabulary image segmentation with image-level labels},
  author={Ghiasi, Golnaz and Gu, Xiuye and Cui, Yin and Lin, Tsung-Yi},
  booktitle={European Conference on Computer Vision},
  pages={540--557},
  year={2022},
  organization={Springer}
}

@inproceedings{ding2022decoupling,
  title={Decoupling zero-shot semantic segmentation},
  author={Ding, Jian and Xue, Nan and Xia, Gui-Song and Dai, Dengxin},
  booktitle={Proceedings of the IEEE/CVF Conference on Computer Vision and Pattern Recognition},
  pages={11583--11592},
  year={2022}
}

@inproceedings{zhou2022extract,
  title={Extract free dense labels from clip},
  author={Zhou, Chong and Loy, Chen Change and Dai, Bo},
  booktitle={European Conference on Computer Vision},
  pages={696--712},
  year={2022},
  organization={Springer}
}

@article{shi2024lca,
  title={LCA-on-the-Line: Benchmarking Out-of-Distribution Generalization with Class Taxonomies},
  author={Shi, Jia and Gare, Gautam and Tian, Jinjin and Chai, Siqi and Lin, Zhiqiu and Vasudevan, Arun and Feng, Di and Ferroni, Francesco and Kong, Shu},
  journal={arXiv preprint arXiv:2407.16067},
  year={2024}
}

@inproceedings{cho2024cat,
  title={Cat-seg: Cost aggregation for open-vocabulary semantic segmentation},
  author={Cho, Seokju and Shin, Heeseong and Hong, Sunghwan and Arnab, Anurag and Seo, Paul Hongsuck and Kim, Seungryong},
  booktitle={Proceedings of the IEEE/CVF Conference on Computer Vision and Pattern Recognition},
  pages={4113--4123},
  year={2024}
}

@inproceedings{radford2021learning,
  title={Learning transferable visual models from natural language supervision},
  author={Radford, Alec and Kim, Jong Wook and Hallacy, Chris and Ramesh, Aditya and Goh, Gabriel and Agarwal, Sandhini and Sastry, Girish and Askell, Amanda and Mishkin, Pamela and Clark, Jack and others},
  booktitle={International conference on machine learning},
  pages={8748--8763},
  year={2021},
  organization={PMLR}
}

@inproceedings{caesar2018coco,
  title={Coco-stuff: Thing and stuff classes in context},
  author={Caesar, Holger and Uijlings, Jasper and Ferrari, Vittorio},
  booktitle={Proceedings of the IEEE conference on computer vision and pattern recognition},
  pages={1209--1218},
  year={2018}
}

@inproceedings{liang2023open,
  title={Open-vocabulary semantic segmentation with mask-adapted clip},
  author={Liang, Feng and Wu, Bichen and Dai, Xiaoliang and Li, Kunpeng and Zhao, Yinan and Zhang, Hang and Zhang, Peizhao and Vajda, Peter and Marculescu, Diana},
  booktitle={Proceedings of the IEEE/CVF Conference on Computer Vision and Pattern Recognition},
  pages={7061--7070},
  year={2023}
}

@article{blumenstiel2024mess,
  title={What a mess: Multi-domain evaluation of zero-shot semantic segmentation},
  author={Blumenstiel, Benedikt and Jakubik, Johannes and K{\"u}hne, Hilde and V{\"o}ssing, Michael},
  journal={Advances in Neural Information Processing Systems},
  volume={36},
  year={2024}
}

@inproceedings{xu2023open,
  title={Open-vocabulary panoptic segmentation with text-to-image diffusion models},
  author={Xu, Jiarui and Liu, Sifei and Vahdat, Arash and Byeon, Wonmin and Wang, Xiaolong and De Mello, Shalini},
  booktitle={Proceedings of the IEEE/CVF Conference on Computer Vision and Pattern Recognition},
  pages={2955--2966},
  year={2023}
}

@inproceedings{xu2022simple,
  title={A simple baseline for open-vocabulary semantic segmentation with pre-trained vision-language model},
  author={Xu, Mengde and Zhang, Zheng and Wei, Fangyun and Lin, Yutong and Cao, Yue and Hu, Han and Bai, Xiang},
  booktitle={European Conference on Computer Vision},
  pages={736--753},
  year={2022},
  organization={Springer}
}

@article{yu2023convolutions,
  title={Convolutions die hard: Open-vocabulary segmentation with single frozen convolutional clip},
  author={Yu, Qihang and He, Ju and Deng, Xueqing and Shen, Xiaohui and Chen, Liang-Chieh},
  journal={Advances in Neural Information Processing Systems},
  volume={36},
  pages={32215--32234},
  year={2023}
}

@article{ding2022open,
  title={Open-Vocabulary Panoptic Segmentation MaskCLIP},
  author={Ding, Zheng and Wang, Jieke and Tu, Zhuowen},
  year={2022}
}

@inproceedings{xu2023side,
  title={Side adapter network for open-vocabulary semantic segmentation},
  author={Xu, Mengde and Zhang, Zheng and Wei, Fangyun and Hu, Han and Bai, Xiang},
  booktitle={Proceedings of the IEEE/CVF Conference on Computer Vision and Pattern Recognition},
  pages={2945--2954},
  year={2023}
}

@article{zhou2019semantic,
  title={Semantic understanding of scenes through the ade20k dataset},
  author={Zhou, Bolei and Zhao, Hang and Puig, Xavier and Xiao, Tete and Fidler, Sanja and Barriuso, Adela and Torralba, Antonio},
  journal={International Journal of Computer Vision},
  volume={127},
  pages={302--321},
  year={2019},
  publisher={Springer}
}

@article{everingham2015pascal,
  title={The pascal visual object classes challenge: A retrospective},
  author={Everingham, Mark and Eslami, SM Ali and Van Gool, Luc and Williams, Christopher KI and Winn, John and Zisserman, Andrew},
  journal={International journal of computer vision},
  volume={111},
  pages={98--136},
  year={2015},
  publisher={Springer}
}

@inproceedings{sakaridis2021acdc,
  title={ACDC: The adverse conditions dataset with correspondences for semantic driving scene understanding},
  author={Sakaridis, Christos and Dai, Dengxin and Van Gool, Luc},
  booktitle={Proceedings of the IEEE/CVF International Conference on Computer Vision},
  pages={10765--10775},
  year={2021}
}

@inproceedings{wortsman2022robust,
  title={Robust fine-tuning of zero-shot models},
  author={Wortsman, Mitchell and Ilharco, Gabriel and Kim, Jong Wook and Li, Mike and Kornblith, Simon and Roelofs, Rebecca and Lopes, Raphael Gontijo and Hajishirzi, Hannaneh and Farhadi, Ali and Namkoong, Hongseok and others},
  booktitle={Proceedings of the IEEE/CVF conference on computer vision and pattern recognition},
  pages={7959--7971},
  year={2022}
}

@article{kumar2022fine,
  title={Fine-tuning can distort pretrained features and underperform out-of-distribution},
  author={Kumar, Ananya and Raghunathan, Aditi and Jones, Robbie and Ma, Tengyu and Liang, Percy},
  journal={arXiv preprint arXiv:2202.10054},
  year={2022}
}

@inproceedings{zou2023generalized,
  title={Generalized decoding for pixel, image, and language},
  author={Zou, Xueyan and Dou, Zi-Yi and Yang, Jianwei and Gan, Zhe and Li, Linjie and Li, Chunyuan and Dai, Xiyang and Behl, Harkirat and Wang, Jianfeng and Yuan, Lu and others},
  booktitle={Proceedings of the IEEE/CVF conference on computer vision and pattern recognition},
  pages={15116--15127},
  year={2023}
}

@article{liang2022holistic,
  title={Holistic evaluation of language models},
  author={Liang, Percy and Bommasani, Rishi and Lee, Tony and Tsipras, Dimitris and Soylu, Dilara and Yasunaga, Michihiro and Zhang, Yian and Narayanan, Deepak and Wu, Yuhuai and Kumar, Ananya and others},
  journal={arXiv preprint arXiv:2211.09110},
  year={2022}
}

@article{lyu2020uavid,
  title={UAVid: A semantic segmentation dataset for UAV imagery},
  author={Lyu, Ye and Vosselman, George and Xia, Gui-Song and Yilmaz, Alper and Yang, Michael Ying},
  journal={ISPRS journal of photogrammetry and remote sensing},
  volume={165},
  pages={108--119},
  year={2020},
  publisher={Elsevier}
}

@misc{swissphoto2012isprs,
  title={ISPRS Potsdam dataset within the ISPRS test project on urban classification, 3D building reconstruction and semantic labeling},
  author={Swissphoto, BSF},
  year={2012}
}

@inproceedings{lee2024caltech,
  title={Caltech aerial rgb-thermal dataset in the wild},
  author={Lee, Connor and Anderson, Matthew and Ranganathan, Nikhil and Zuo, Xingxing and Do, Kevin and Gkioxari, Georgia and Chung, Soon-Jo},
  booktitle={European Conference on Computer Vision},
  pages={236--256},
  year={2024},
  organization={Springer}
}

@inproceedings{ji2023multispectral,
  title={Multispectral video semantic segmentation: A benchmark dataset and baseline},
  author={Ji, Wei and Li, Jingjing and Bian, Cheng and Zhou, Zongwei and Zhao, Jiaying and Yuille, Alan L and Cheng, Li},
  booktitle={Proceedings of the IEEE/CVF Conference on Computer Vision and Pattern Recognition},
  pages={1094--1104},
  year={2023}
}

@article{tang2022piafusion,
  title={PIAFusion: A progressive infrared and visible image fusion network based on illumination aware},
  author={Tang, Linfeng and Yuan, Jiteng and Zhang, Hao and Jiang, Xingyu and Ma, Jiayi},
  journal={Information Fusion},
  volume={83},
  pages={79--92},
  year={2022},
  publisher={Elsevier}
}

@article{sikdar2025saga,
  title={SAGA: Semantic-Aware Gray color Augmentation for Visible-to-Thermal Domain Adaptation across Multi-View Drone and Ground-Based Vision Systems},
  author={Sikdar, Aniruddh and Gurunath, Prajwal and Udupa, Sumanth and Sundaram, Suresh and others},
  journal={arXiv preprint arXiv:2504.15728},
  year={2025}
}

@inproceedings{paul2022vision,
  title={Vision transformers are robust learners},
  author={Paul, Sayak and Chen, Pin-Yu},
  booktitle={Proceedings of the AAAI conference on Artificial Intelligence},
  volume={36},
  number={2},
  pages={2071--2081},
  year={2022}
}

@inproceedings{seifi2021glimpse,
  title={Glimpse-attend-and-explore: Self-attention for active visual exploration},
  author={Seifi, Soroush and Jha, Abhishek and Tuytelaars, Tinne},
  booktitle={Proceedings of the IEEE/CVF International Conference on Computer Vision},
  pages={16137--16146},
  year={2021}
}

@inproceedings{ji2023semanticrt,
  title={Semanticrt: A large-scale dataset and method for robust semantic segmentation in multispectral images},
  author={Ji, Wei and Li, Jingjing and Bian, Cheng and Zhang, Zhicheng and Cheng, Li},
  booktitle={Proceedings of the 31st ACM International Conference on Multimedia},
  pages={3307--3316},
  year={2023}
}

@article{yuan2024unirgb,
  title={UniRGB-IR: A Unified Framework for Visible-Infrared Semantic Tasks via Adapter Tuning},
  author={Yuan, Maoxun and Cui, Bo and Zhao, Tianyi and Wang, Jiayi and Fu, Shan and Yang, Xue and Wei, Xingxing},
  journal={arXiv preprint arXiv:2404.17360},
  year={2024}
}

@inproceedings{franchi2024infraparis,
  title={Infraparis: A multi-modal and multi-task autonomous driving dataset},
  author={Franchi, Gianni and Hariat, Marwane and Yu, Xuanlong and Belkhir, Nacim and Manzanera, Antoine and Filliat, David},
  booktitle={Proceedings of the IEEE/CVF Winter Conference on Applications of Computer Vision},
  pages={2973--2983},
  year={2024}
}

@article{portales2023global,
  title={Global flood extent segmentation in optical satellite images},
  author={Portal{\'e}s-Juli{\`a}, Enrique and Mateo-Garc{\'\i}a, Gonzalo and Purcell, Cormac and G{\'o}mez-Chova, Luis},
  journal={Scientific reports},
  volume={13},
  number={1},
  pages={20316},
  year={2023},
  publisher={Nature Publishing Group UK London}
}

@inproceedings{xia2023openearthmap,
  title={Openearthmap: A benchmark dataset for global high-resolution land cover mapping},
  author={Xia, Junshi and Yokoya, Naoto and Adriano, Bruno and Broni-Bediako, Clifford},
  booktitle={Proceedings of the IEEE/CVF Winter Conference on Applications of Computer Vision},
  pages={6254--6264},
  year={2023}
}

@article{wang2021loveda,
  title={LoveDA: A remote sensing land-cover dataset for domain adaptive semantic segmentation},
  author={Wang, Junjue and Zheng, Zhuo and Ma, Ailong and Lu, Xiaoyan and Zhong, Yanfei},
  journal={arXiv preprint arXiv:2110.08733},
  year={2021}
}

@article{kendall2017uncertainties,
  title={What uncertainties do we need in bayesian deep learning for computer vision?},
  author={Kendall, Alex and Gal, Yarin},
  journal={Advances in neural information processing systems},
  volume={30},
  year={2017}
}

@inproceedings{zhai2023sigmoid,
  title={Sigmoid loss for language image pre-training},
  author={Zhai, Xiaohua and Mustafa, Basil and Kolesnikov, Alexander and Beyer, Lucas},
  booktitle={Proceedings of the IEEE/CVF international conference on computer vision},
  pages={11975--11986},
  year={2023}
}

@article{xie2025fg,
  title={FG-CLIP: Fine-Grained Visual and Textual Alignment},
  author={Xie, Chunyu and Wang, Bin and Kong, Fanjing and Li, Jincheng and Liang, Dawei and Zhang, Gengshen and Leng, Dawei and Yin, Yuhui},
  journal={arXiv preprint arXiv:2505.05071},
  year={2025}
}

@inproceedings{sikdar2025picazo,
  title={Picazo: Pixel-aligned contrastive learning for zero-shot domain adaptation},
  author={Sikdar, Aniruddh and Kishor, Arya and Kadam, Ishika and Sundaram, Suresh},
  booktitle={Proceedings of the IEEE/CVF Conference on Computer Vision and Pattern Recognition},
  pages={6600--6610},
  year={2025}
}

@article{gandhamal2025ov,
  title={Ov-coast: Cost aggregation with optimal transport for open-vocabulary semantic segmentation},
  author={Gandhamal, Aditya and Sikdar, Aniruddh and Sundaram, Suresh},
  journal={arXiv preprint arXiv:2506.03706},
  year={2025}
}

@inproceedings{chaudhuri2025domain,
  title={Domain generalization for environmental uncertainties using Metacognitive Mask R-CNN},
  author={Chaudhuri, Sayan and Sikdar, Aniruddh and Sundaram, Suresh},
  booktitle={2025 International Conference on Emerging Technology in Autonomous Aerial Vehicles (ETAAV)},
  pages={1--6},
  year={2025},
  organization={IEEE}
}
\end{document}